# CONSTRUCTING THE PIGNISTIC PROBABILITY FUNCTION IN A CONTEXT OF UNCERTAINTY.

Philippe SMETS[1]
I.R.I.D.I.A. Université Libre de Bruxelles
50 av. Roosevelt, CP 194-6.
B-1050, Brussels, Belgium. (tel.:+322.642.2729)

**Summary:** We derive axiomatically the probability function that should be used to make decisions given **any** form of underlying uncertainty.

## 1. Introduction.

### 1.1. Beliefs.

Many new models have been proposed to quantify uncertainty. But usually they don't explain how decisions must be derived. In probability theory, the expected utility model is well established and strongly justified. We show that such expected utility model can be derived in the other models proposed to quantify someone's belief. The justification is based on special bets and some coherence requirements that lead to the derivation of the so-called **generalized insufficient reason principle**. In Smets (1988b, 1988c, 1989) we emphasize the existence of two levels where beliefs manifest themselves: the **credal level** where beliefs are entertained and the **pignistic level** where beliefs are used to take decisions (pignus = a bet in Latin, Smith 1961).

Uncertainty induces beliefs, i.e. graded dispositions that guide our behaviour. Translated within a normative approach, this leads usually to the construction of a model to quantify beliefs that is linked directly to "rational" agent behaviour within betting and decision contexts (DeGroot 1970). Bayesians have convincingly showed that if decisions must be "coherent", our belief over the various outcomes must be quantified by a probability function. This result is accepted here. Hence at the pinistic level beliefs are quantified by probability functions. But *probability functions are used to quantify our belief only when a decision is really involved.* That beliefs are necessary ingredients for our decisions does not mean that beliefs cannot be entertained without any revealing behaviour manifestations (Smith and Jones, p.147).

The probability function that quantifies our belief at the pignistic level reflects an underlying credal state. At the credal level, quantified beliefs can be represented by other models like the transferable belief model (Smets 1988a, 1989a), the possibility theory model (Dubois and Prade 1988), the upper and lower probabilities models Good (1950), convex sets of probability functions (Kyburg 1987b), Dempster-Shafer's models (Shafer 1976), etc... The difference between these models are studied in detail in Smets (1989a, 1989b). The major difference results from the existence or the absence of an underlying probability function.

In Smets (1989a) we present the semantic of belief functions within the transferable belief model. It is based on betting behaviors and the consideration of varying betting frames. One should notice that the transferable belief model does not correspond to the Dempster-Shafer model as used in AI, but is very close to what Shafer developed in his book (Shafer 1976).

All beliefs entertained by an agent are defined relative to a given **doxastic corpus** that consists of all pieces of information in the agent's possession. Our approach is normative: the agent is an ideal rational subject and the doxastic corpus is deductively closed.

---

The function that quantifies our belief at the credal level will be called here a **credibility function**, denoted Cr. We introduce such name as a generic for the familiy of functions proposed to quantify beliefs like the probability functions, the belief functions, the lower probabilities functions, the necessity functions and their dual.

Our derivation of the pignistic probability function fits with any credibility function. It is not restricted to any of the above mentioned functions. It fits with Dempster-Shafer model, but is not restricted to it. Hence the "credibility" qualification used to enhance the generality of our results.

Beliefs being the governing principles of our decisions, the (pignistic) probability functions observed at the pignistic level must be derived from the credibility functions present at the credal level. Some forms of coherence must be satisfied by this transformation. These coherence requirements lead to the derivation of an unique transformation. This paper presents these requirements and the derived transformation that turns out to be an application of the generalized insufficient reason principle (Dubois and Prade 1982, Williams 1982, Smets 1988a)

### 1.2. The propositional space.

Let $\Omega$ be a non empty finite set called the frame of discernment equipped with a Boolean algebra $\Re$ of some of its subsets. Every element of $\Re$ is called a proposition. The pair $(\Omega, \Re)$ is called a **propositional space**. By abuse of language, the elements of $\Omega$ are called the **elementary propositions**. Any algebra $\Re$ defined on $\Omega$ contains two special propositions: $1_\Omega$ and $0_\Omega$ where $1_\Omega$ is the set of all elementary propositions of $\Omega$ and $0_\Omega$ is the complement of $1_\Omega$ relative to $\Omega$. The standard Boolean notation is used for the propositions of $\Re$: $\overline{A}$ stands for the complement of A relative to $\Omega$, and $A \cup B$, $A \cap B$ denotes the set-theoretic union and intersection of propositions A and B of $\Re$. $A \subseteq B$ means that all elementary propositions of A are elementary propositions of B. Each proposition in $\Re$, except $0_\Omega$, such that its intersection with any proposition in $\Re$ is either itself or $0_\Omega$ is called an atom of $\Re$. Every proposition in $\Re$ can be described as the union of **atoms** of $\Re$.

On $\Omega$, we define a valuation that maps every element of $\Omega$ into {true, false} such that at most one element of $\Omega$ is true; this element is called the 'truth'. A proposition in $\Omega$ is true iff one of its elementary propositions is true.

Two propositions A and B are doxastically **equivalent** (or simply equivalent) if they share the same truth value for all valuations that satisfy the doxastic corpus constraints: it is denoted $A \equiv B$.

### 2. The credibility function.

Suppose an agent, with his/her doxastic corpus, entertains beliefs over a frame $\Omega$, i.e. he/she assigns degrees of belief to the elements of $\Re$, an algebra defined on $\Omega$. It is postulated that degrees of belief are quantified by a point-valued "credibility" function Cr which maps $\Re$ into a closed interval of the real line, is monotonic for inclusion, reaches its lower limit for $0_\Omega$ and its upper limit for $1_\Omega$ if $1_\Omega$ is equivalent to a tautology $\tau$. (That $1_\Omega$ might be different from a tautology reflects the distinction between the open-world and the closed-world assumptions as explained in Smets (1988a).

**Axiom A1**: Let a propositional space $(\Omega, \Re)$. Given a doxastic corpus, there exists a unique function Cr that quantifies the agent's belief for every proposition of $\Re$.

The triple $(\Omega, \Re, Cr)$ is called a **credibility space** and denoted by $\wp$. The index i in $\wp_i$ corresponds to the index of $(\Omega_i, \Re_i, Cr_i)$

**Axiom A2: Domain.**
Given a credibility space $\wp$, $Cr: \Re \to [\phi, \psi]$ where $[\phi, \psi]$ is an interval of the real line.

**Axiom A3: Monotonicity.**
Given a credibility space $\wp$, $\forall A, B \in \Re$, if $A \subseteq B$, then $Cr(A) \leq Cr(B)$

**Axiom A4: Lower limit.**
Given a credibility space $\wp$, $Cr(0_\Omega) = \phi$

**Axiom A5: Upper limit.**
Given a credibility space $\wp$, if $1_\Omega \equiv \tau$, then $Cr(1_\Omega) = \psi$

Doxastically equivalent propositions should always receive equal credibilities (Kyburg, 1987a)

**Axiom A6: equi-credibility of doxastically equivalent propositions.**
Suppose two credibility spaces $(\Omega_i, \Re_i, Cr_i)$, $i=1,2$. If $A_1 \in \Re_1$, $A_2 \in \Re_2$ and $A_1 \equiv A_2$, then $Cr_1(A_1) = Cr_2(A_2)$.

Suppose two credibility spaces $(\Omega, \Re_i, Cr_i)$, $i=1,2$, defined on the same space $\Omega$. Axiom A6 implies that those propositions that belong to both algebras will always receive the same credibility as they are logically equivalent. Hence the credibility given to a proposition does not depend on the structure of the algebra to which the proposition belongs.

Axiom A6 permits to prove the following anonymity theorem.

**Theorem 1:** Let G be a permutation function defined on $\Omega$. For $A \subseteq \Omega$, let $G(A) = \{G(x): x \in A\}$. Let a credibility space $(\Omega, \Re, Cr)$. Let $Cr'$ be the credibility function defined on $\Re' = \{G(A): A \in \Re\}$. Then, by axiom A6, $\forall A \in \Re$,
$$Cr'(G(A)) = Cr(A)$$

### 3. $\alpha$-combined credibility spaces.

We are going to show that the set of credibility functions defined on a propositional space $(\Omega, \Re)$ is a convex set. To show this we introduce a **combined bet schema**. Suppose two propositional spaces $(\Omega_i, \Re_i)$ $i=1,2$, where the atoms of $\Re_i$ are $\{A_{i1}, A_{i2}, ..., A_{in}\}$. Let $N = \{1, 2...n\}$. Such a pair of propositional spaces (and the corresponding credibility spaces) are said **combinable**.

Let $\wp_i$, $i=1,2$ be two combinable credibility spaces. Suppose a random generator that generates event R with $P(R=r)=\alpha$ and $P(R=s)=1-\alpha$. Define the $\alpha$-**combined credibility space** $\wp_{12} = (\Omega_{12}, \Re_{12}, Cr_{12})$ with atoms $\{A_{12j}: j = 1, 2...n\}$. The valuation on $\Omega_{12}$ is such that $A_{12j}$ is true if (r occurs and $A_{1j}$ is true) or (s occurs and $A_{2j}$ is true). For any $I \subseteq N$, let $A_{iI} = \bigcup_{j \in I} A_{ij}$ and $A_{12I} = \bigcup_{j \in I} A_{12j}$. $A_{iI}$ and $A_{12I}$ are the propositions of the algebras $\Re_i$ and $\Re_{12}$. $A_{iI}$ is true if one of the propositions $A_{ij}: j \in I$, is true. $A_{12I}$ is true if (r occurs and $A_{1I}$ is true) or (s occurs and $A_{2I}$ is true).

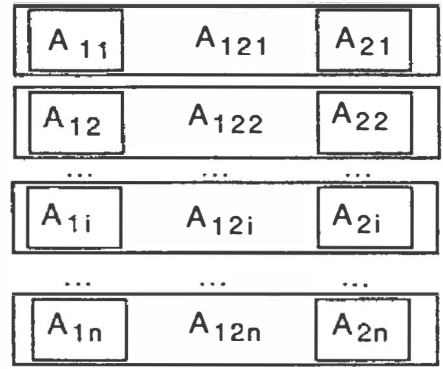

**Figure 1:** Two combinable credibility spaces with atoms $A_{1i}$ and $A_{2i}$, $i=1,...n$, and the $\alpha$ combined credibility space with atoms $A_{12i}$. Atom $A_{12i}$ is true whenever (R=r and $A_{1i}$ is true) or (R=s and $A_{2i}$ is true) where $P(R=r)=\alpha$ and $P(R=s)=1-\alpha$.

We will show in §4 that under natural requirements, one has:
$Cr_{12}(A_{12I}) = \alpha\, Cr_1(A_{1I}) + (1-\alpha)\, Cr_2(A_{2I})$

Suppose two combinable credibility spaces $\wp_i$, $i=1,2$, and their associated $\alpha$-combined credibility space $\wp_{12}$. The following axioms are postulated for credibility function $Cr_{12}$.

**Axiom C1: pointwise compositionality.**
There exists a function $F:[\phi, \psi]^2 \to [\phi, \psi]$ such that $\forall I \subseteq N$
$$Cr_{12}(A_{12I}) = F(Cr_1(A_{1I}), Cr_2(A_{2I}))$$

**Axiom C2: continuity.**
$F(x,y)$ is continuous in $(x,y) \in [\phi, \psi]^2$

**Axiom C3: strict monotony.**
$F(x,y)$ is strictly monotonic for $x, y \in [\phi, \psi]$

**Axiom C4: idempotency.**
$F(x,x) = x \quad \forall x \in [\phi, \psi]$

Pointwise compositionality is justified by the idea that $Cr_{12}(A_{12I})$ should not be changed if we replace $\Re_i$ by the algebra $\Re_i'$ whose only two atoms are $A_{iI}$ and $A_{iI'}$ where $I'$ is the complement of $I$ relative to $N$ (with parallel definitions for $\Re_{12}$).

Continuity is classically accepted, it could be weakened but without real profit.

Strict monotony is postulated as we consider that $Cr_{12}$ should be sensitive to both of its components. The credibility of one proposition in $\Re_1$ should not inhibited even locally the impact of our credibility on another proposition in $\Re_2$ (and vice versa).

Finally idempotency reflects the idea that if $\wp_1$ and $\wp_2$ happen to be the same credibility spaces, then $Cr_{12} = Cr_1 = Cr_2$.

In theorem 2, we show that F satisfies the bisymmetry equation
$F( F(x,y) , F(u,v) ) = F( F(x,u) , F(y,v) )$
whose solution is detailed in Aczel (1966, pg. 287)

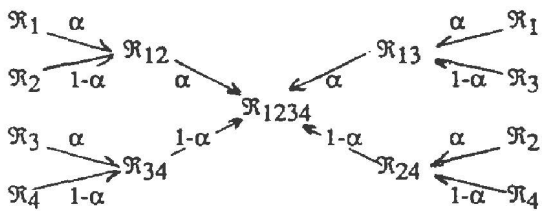

**Figure 1**: creation in the proof of theorem 2 of $\Re_{1234}$ by combining $\Re_{12}$ with $\Re_{34}$, or $\Re_{13}$ with $\Re_{24}$.

**Theorem 2**: Given two combinable credibility spaces $\wp_i$, $i=1,2$, and their associated $\alpha$-combined credibility space $\wp_{12}$, given axioms C1 to C4, then the F function in axiom C1 satisfies:

$F(x,y) = f( a.f^{-1}(x) + (1-a).f^{-1}(y) )$ (1)

where $f(x) \in [\phi,\psi]$, is continuous, strictly monotonic, $f(\phi) = \phi$ and $f(\psi) = \psi$, and $a \in [0, 1]$.

**Proof**: Consider four combinable credibility spaces $\wp_i$, $i=1,4$, where the atoms of $\Re_i$ are $\{A_{i1},... A_{in}\}$. Build the $\alpha$-combined credibility spaces $\wp_{12}$, $\wp_{34}$, $\wp_{13}$ and $\wp_{24}$.

By axiom C1, $\forall I \subseteq N$, $i,j \in \{1, 2, 3, 4\}$, $Cr_{ij}(A_{ijI}) = F( Cr_i(A_{iI}) , Cr_j(A_{jI}) )$

Consider also the join $\alpha$-combined credibility space $\wp(12)(34) = (\Omega_{(12)(34)}, \Re_{(12)(34)}, Cr_{(12)(34)})$ build from $\wp_{12}$ and $\wp_{34}$. Let the result from successive uses of the R-device be stochastically independent. Let $R_1$, $R_2$ and $R_3$ be the three independent random variables generated by the R-device that will be used respectively to select between $\wp_{12}$ and $\wp_{34}$, between $\wp_1$ and $\wp_2$, between $\wp_3$ and $\wp_4$. Let $R_i \in \{r_i, s_i\}$ with $P(R_i=r_i) = \alpha$ for $i=1, 2, 3$. Proposition $A_{(12)(34)j}$ in $\Re_{(12)(34)}$ is true if ($r_1$ and $r_2$ occur and $A_{1j}$ is true) or ($r_1$ and $s_2$ and $A_{2j}$) or ($s_1$ and $r_3$ and $A_{3j}$) or ($s_1$ and $s_3$ and $A_{4j}$).

Consider then the join $\alpha$-combined credibility space $\wp(13)(24)$ build from $\wp_{13}$ and $\wp_{24}$. Proposition $A_{(13)(24)j}$ in $\Re_{(13)(24)}$ is true if ($r_1$ and $r_2$ and $A_{1j}$) or ($r_1$ and $s_2$ and $A_{3j}$) or ($s_1$ and $r_3$ and $A_{2j}$) or ($s_1$ and $s_3$ and $A_{4j}$).

Hence to decide that propositions $A_{(12)(34)j}$ and $A_{(13)(24)j}$ are true, one check the propositions within $\Re_i$ where the index i is selected by a random process, each value having the same chance to be selected in the two join $\alpha$-combined credibility spaces $\wp(12)(34)$ and $\wp(13)(24)$: $P(i=1) = \alpha^2$, $P(i=2) = P(i=3) = \alpha(1-\alpha)$ and $P(i=4) = (1-\alpha)^2$

This identity means that the credibility functions $Cr_{(12)(34)}$ and $Cr_{(13)(24)}$ are identical, hence the bisymmetry equation: $\forall I \subseteq N$
$F(F( Cr_1(A_{1I}) , Cr_2(A_{2I}) ) , F( Cr_3(A_{3I}) , Cr_4(A_{4I})))$
$=F(F( Cr_1(A_{1I}) , Cr_3(A_{3I}) ) , F( Cr_2(A_{2I}) , Cr_4(A_{4I})))$

By axioms C2 and C3, F is continuous and strictly increasing in both variables $x,y \in [\phi,\omega]$, hence its general solution is

$F(x,y) = f( a.f^{-1}(x) + b.f^{-1}(y) + c )$

where $f(x) \in [\phi,\psi]$, is continuous and strictly monotone (see theorem 1, page 287, Aczel 1966). By axiom C4, $F(x,x) = x$, then $c = 0$ and $a + b = 1$. Such F function satisfies axioms A4 and A5 with $f(\phi) = \phi$ and $f(\psi) = \psi$. QED

## 4. The pignistic probability function.

Suppose we have a credibility space $\wp = (\Omega, \Re, Cr)$. When a decision must be taken that depends on the proposition in $\Re$ that will be true, one must construct a pignistic probability function on $\Re$ in order to take the optimal decision that maximizes the expected utility. We assume, as explained in the introduction, that the pignistic probability function defined on $\Re$ is a function of the credibility function Cr. Hence one must transform Cr into a probability function P. This transformation is called hereafter the $\Gamma_\Re$ transformation where the $\Re$ index mentions the Boolean algebra on which Cr and P are defined: so $P = \Gamma_\Re(Cr)$. It is also postulated that the transformation depends only on the cardinality of $\Re$, not on the nature of its atoms.

**Axiom P1**: Let a credibility space $\wp = (\Omega, \Re, Cr)$ and $P = \Gamma_\Re(Cr)$. For any atom $\omega$ of $\Re$,
$$P(\omega) = g(\omega, \{Cr(A): A \in \Re\})$$

Axiom P1 formalizes the idea that our beliefs guide our behaviours. Evaluation of P for non atoms of $\Re$ is obtained by adding the appropriate probabilities.

**Axiom P2**: Suppose two combinable credibility spaces $\wp_1$ and $\wp_2$, and their associated $\alpha$-combined credibility space $\wp_{12}$. Let $P_1 = \Gamma_{\Re_1}(Cr_1)$, $P_2 = \Gamma_{\Re_2}(Cr_2)$ and $P_{12} = \Gamma_{\Re_{12}}(Cr_{12})$. Let $A_{1j}$, $A_{2j}$ and $A_{12j}$ $j \in N = \{1, 2, \ldots n\}$, be the atoms of the three algebras. Let $A_{1I} = \bigcup_{j \in I} A_{1j}$, $A_{2I} = \bigcup_{j \in I} A_{2j}$ and $A_{12I} = \bigcup_{j \in I} A_{12j}$ where $I \subseteq N$. Then, $\forall I \subseteq N$,
$$P_{12}(A_{12I}) = \alpha . P_1(A_{1I}) + (1-\alpha) . P_2(A_{2I})$$

Axiom P2 formalizes in the present context the well-known property:
$$P(X) = P(X|A).P(A) + P(X|\overline{A}).P(\overline{A})$$
as the $P_i(A_{iI})$ are the conditional probabilities $P_{12}$ of $A_{12I}$ in context $\Re_i$ and $\alpha$ is the probability of the context $\Re_1$. e.g. $P_1(A_{1I}) = P_{12}(A_{12I}|R=r)$.

Axiom P2 implies that the function f in theorem 2 is such that $f(x) = x$.

Indeed suppose $Cr_i$ are probability functions $P_i$, then $Cr_{12}$ is also a probability function $P_{12}$ with:
$Cr_{12}(A_{12I}) = P_{12}(A_{12I}) = P_{12}(A_{12I}|R=r) \alpha + P_{12}(A_{12I}|R=s) (1-\alpha)$
As $P_{12}(A_{12I}|R=r) = P_1(A_{1I})$ and $P_{12}(A_{12I}|R=s) = P_2(A_{2I})$, (1) becomes:
$P_{12}(A_{12I}) = F(P_1(A_{1I}), P_2(A_{2I})) = f(af^{-1}(P_1(A_{1I})) + (1-a)f^{-1}(P_2(A_{2I}))) = \alpha P_1(A_{1I}) + (1-\alpha)P_2(A_{2I})$

i.e. $f(af^{-1}(x) + (1-a)f^{-1}(y)) = \alpha x + (1-\alpha) y$
what implies that $f^{-1}(x) = x$ and $a = \alpha$.

Therefore (1) becomes:
$Cr_{12}(A_{12I}) = \alpha Cr_1(A_{1I}) + (1-\alpha) Cr_2(A_{2I})$ (2)

The anonymity property of theorem 1 is generalized to pignistic probabilities.

**Axiom P3: anonymity**: Let G be a permutation function defined on $\Omega$. For $A \subseteq \Omega$, let $G(A) = \{G(x): x \in A\}$. Let a credibility space $(\Omega, \Re, Cr)$ and $P = \Gamma_\Re$. Let the credibility space $(\Omega, \Re', Cr')$ where $\Re' = \{G(A): A \in \Re\}$. Let $P' = \Gamma_{\Re'}$. Then $\forall A \in \Re$,
$$P'(G(A)) = P(A).$$

Let a credibility space $\wp = (\Omega, \Re, Cr)$ and $P = \Gamma_\Re(Cr)$. As far as P is a probability function, it must satisfy the following obvious properties:

**Axiom Q1: Sum property**: If $Cr(1_\Omega) = \psi$, then $P(1_\Omega) = 1$

**Axiom Q2: False event**: If $X \in \Re$ and $X \equiv \emptyset$, then $Cr(A) = Cr(A \cup X)$ $\forall A \in \Re$ and $P(X) = 0$. ($\emptyset$ is the logical contradiction)

**Axiom Q3: Credibility = Probability**: If Cr happens to be a probability function P defined on $\Re$, then $\Gamma_\Re(P) = P$.

Axiom Q1 tells that if $\Re$ is rich enough so that $1_\Omega$ is equivalent to a tautology, then the probabilities given to the atoms of $\Re$ add to one.

Axiom Q2 tells that the credibility given to a proposition is not changed when one adds any impossible proposition to it.

Axioms Q3 recognizes that if someone's credibility is already described by a probability function, then the pignistic probabilities and the credibilities are equal.

Theorem 3 shows that the g function of axiom P1 is a linear function of its arguments $Cr(A): A \in \Re$. Figure 3 illustrates the origin of this linearity. There are two ways by which $P_{12}$ can be constructed: either build each $P_i$ and combine them or combine both $Cr_i$ into

323

$Cr_{12}$ and transform the last into $P_{12}$. Both approaches must give the same answer.

Axioms Q1 to Q3 permit then to derive all the coefficients of (3).

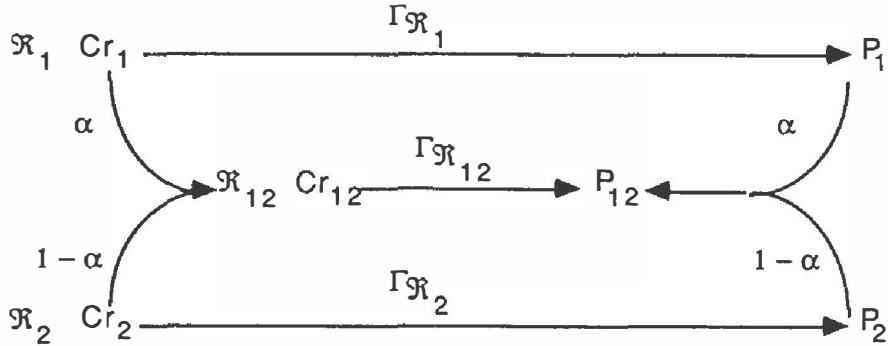

**Figure 3:** Graphical representation of the linearity of $\Gamma$.
$$\Gamma_{\Re_{12}}(\alpha Cr_1 + (1-\alpha)Cr_2) = \alpha \Gamma_{\Re_1}(Cr_1) + (1-\alpha)\Gamma_{\Re_2}(Cr_2)$$

**Theorem 3.** Given a credibility space $(\Omega, \Re, Cr)$ and $P = \Gamma_\Re(Cr)$. Given (2), axioms P1 and P2, then there exists a and b such that for any atom $\omega$ of $\Re$,
$$P(\omega) = \sum_{A \in \Re} a(\omega, A).Cr(A) + b(\omega) \quad (3)$$

**Proof.** Let $P(R=r) = \alpha$, $\beta = 1-\alpha$, $x_I = Cr_1(A_{1I})$ and $y_I = Cr_2(A_{2I})$. (2) becomes: $\forall I \subseteq N$
$$Cr_{12}(A_{12I}) = \alpha.x_I + \beta.y_I$$
Replacing P in axiom P2, one has: (as $\omega$ is fixed, it is dropped from the notation)
$$g(\alpha.x_1+\beta.y_1, \alpha.x_2+\beta.y_2,...) = \alpha.g(x_1, x_2...) + \beta.g(y_1, y_2...)$$
The proof is based on the transformation of this relation into a Pexider equation (Aczel, 1966, p. 141) and then a Cauchy equation (Aczel, 1966, p. 214). Hence g is linear in its components and the coefficients may depend only on I and $\omega$.

QED

As a consequence of the anonymity axiom P3 and theorem 1, it can easily be shown that the coefficients a and b in (3) depend only on the number of atoms in A and $\omega \cap A$.

**Theorem 4.** Given theorem 3, theorem 1 and axiom P3, relation (3) becomes: $\forall$ atom $\omega$ of $\Omega$
$$P(\omega) = \sum_{A \in \Re} a(|\omega \cap A|, |A|).Cr(A) + b$$
where $|A|$ is the number of atoms of $\Re$ in A.

**Theorem 5**: Given a credibility space $(\Omega, \Re, Cr)$ and $P=\Gamma_\Re(Cr)$. Given theorem 4. Axioms $Q_1$ to $Q_3$ imply:
$$\phi = 0 \qquad \psi = 1$$
and (3) becomes for any atom $\omega_j$ of $\Re$,
$$P(\omega_j) = Cr(\omega_j) + \sum_{i=1}^{n} \frac{1}{n} \frac{1}{\binom{n-1}{i}} \sum_{\substack{j \notin I \\ |I|=i}} (Cr(\omega_j \cup A_I) - Cr(A_I)) \quad (4)$$
where $I \subseteq N = \{1, 2...n\}$

The transformation $\Gamma_\Re$ permits the construction of a probability function (called the pignistic probability function) at the pignistic level given any credibility function at the credal level.

**5. Co-credibility function.**

Given a credibility function Cr on a propositional space $(\Omega, \Re)$, define the co-credibility function CoCr as
$$CoCr(A) = Cr(1_\Omega) - Cr(\bar{A}) \quad \forall A \subseteq \Omega$$
Replacing Cr by its dual CoCr in theorem 3 leads to the same probability function P. For any pair (Cr, CoCr)
$$\Gamma_\Re(Cr) = \Gamma_\Re(CoCr)$$
Using Cr or its dual CoCr is equivalent.

## 6. The Moebius transformations of Cr.

For any credibility space $\wp$, there are two Moebius transforms v and w of the credibility function defined on $\Re$ such that $\forall A \in \Re$ (with all summations taken on those B that are propositions of $\Re$):

$$v(A) = \sum_{B \subseteq A} (-1)^{|A|-|B|} Cr(B)$$

$$v(0_\Omega) = 1 - Cr(1_\Omega)$$

$$Cr(A) = \sum_{\emptyset \neq B \subseteq A} v(B)$$

$$w(A) = \sum_{B \subseteq A} (-1)^{|A|-|B|} (Cr(1_\Omega) - Cr(\overline{B}))$$

$$w(0_\Omega) = 1 - Cr(1_\Omega)$$

$$Cr(A) = \sum_{B \cap A \neq \emptyset} w(B)$$

The transformation between Cr, v and w are one to one. In belief functions theory, the v's are the basic belief masses if Cr is a belief function, and the w's are the basic belief masses if Cr is a plausibility function.

Given the v and w functions, one has:

$$CoCr(A) = \sum_{\emptyset \neq B \subseteq A} w(B)$$

and

$$CoCr(A) = \sum_{B \cap A \neq \emptyset} v(B)$$

Should Cr be respectively a belief, necessity, lower probability or probability function then CoCr would be a plausibility, possibility, upper probability or probability function, and vice versa.

Replacing Cr by v or w, (4) becomes:

$$P(\omega_j) = \sum_{j \in I \subseteq N} v(A_I) / |I| = \sum_{j \in I \subseteq N} w(A_I) / |I|$$

where the $A_I$'s are the propositions of $\Re$.

Hence $P(A) = \sum_{B \subseteq \Omega} v(B) \frac{|A \cap B|}{|B|}$

$$= \sum_{B \subseteq \Omega} w(B) \frac{|A \cap B|}{|B|} \quad \forall A \in \Re$$

a solution that corresponds to the **generalized insufficient reason principle**: for each $B \in \Re$, v(B) (or w(B)) is distributed equally among the atoms of B, and P($\omega$) is the sum of those parts of v (or w) that were given to the atom $\omega$.

## 7. Conclusions.

The generalized insufficient reason principle had already been proposed intuitively as a potential solution to derive a probability function from a belief function (Dubois and Prade 1982, Williams 1982, Smets 1988a) but never justified. We provide an axiomatic justification of this principle based on coherence between combined bets and applicable for any measure of belief whose major property is to be monotonic for set inclusion.

Hence any model for quantified beliefs can be endowed with the needed procedure to transform someone's beliefs entertained at the credal level into a pignistic probability that can be used at the pignistic level when decisions must be taken. This transformation and its justification should answer to the classical criticism of the Dempster-Shafer model and other models based on belief functions, on possibility functions, on upper and lower probabilities functions, etc... Decisions are then based on expected utility theory, using the pignistic probability function to compute the needed expectations.

The link of this model with practical decision problems is straighforward. Given a credibility function Cr that quantifies your degree of belief, if you must make a decision, transform Cr into the pignistic probability function by applying the generalized insufficient reason principle and then use this probability function to select the optimal decision. The whole classical decision theory (the expected utility theory) applies directly (DeGroot 1970, Raiffa 1970)

A final question is: why to bother with a two-level model if decisions are to be based on probability functions as advocated by the Bayesians. What is the need for introducing a credal level and credibility functions. The answer is fully developed in Smets (1989a) where examples are provided that show that the introduction of a two-level model leads to decisions different from those obtained when only one level is considered. Let a first doxastic corpus with the induced credibility function Cr on some propositional space $(\Omega, \Re)$ and the corresponding pignistic probability function $P=\Gamma_\Re(Cr)$. Suppose a new piece of evidence "A is true" with $A \in \Re$ is added to the doxastic corpus. Where to apply the updating that reflects this conditioning on A. We advocate it should be applied at the credal level, by conditioning Cr into $Cr_A$. The pignistic probability $\Gamma_\Re(Cr_A)$ is derived from $Cr_A$. It is usually different from the conditional probability function $P_A$ obtained by conditioning P on A (see Mr. Jones murdering paradigm in Smets, 1988b, 1988c, 1989a).

## Bibliography.

## Acknowledgements.

The author is indebted to R. Kennes, B. Marchal, and A. Saffiotti for their remarks and comments.